\documentclass[letterpaper, 10 pt, conference]{ieeeconf}  

\IEEEoverridecommandlockouts                              

\usepackage{graphicx} 
\usepackage{array}
\usepackage{amsmath}
\usepackage[colorlinks=true, urlcolor=blue, linkcolor=black, citecolor=black]{hyperref}
\usepackage{url}            
\usepackage{amsfonts}       
\usepackage{nicefrac}       
\usepackage{microtype}      
\usepackage{xcolor}         
\usepackage{wrapfig}
\usepackage{booktabs}       
\usepackage{multirow}

\makeatletter

\newcommand{\@IEEEmakecaptiontext}[2]{%
  \setbox\@tempboxa\hbox{\footnotesize #1.~~ #2}%
  \ifdim \wd\@tempboxa >\hsize%
    \setbox\@tempboxa\hbox{\footnotesize #1.~~ }%
    \parbox[t]{\hsize}{%
      \footnotesize
      \noindent\unhbox\@tempboxa#2%
    }%
  \else%
    \ifcenterfigcaptions
      \hbox to\hsize{\footnotesize\hfil\box\@tempboxa\hfil}%
    \else
      \hbox to\hsize{\footnotesize\box\@tempboxa\hfil}%
    \fi
  \fi
}

\long\def\@makecaption#1#2{%
  \ifx\@captype\@IEEEtablestring%
    \@IEEEmakecaptiontext{#1}{#2}%
    \@IEEEtablecaptionsepspace%
  \else
    \@IEEEfigurecaptionsepspace%
    \@IEEEmakecaptiontext{#1}{#2}%
  \fi
}

\makeatother

\title{\LARGE \bf
EgoPhys: Physics-Informed Digital Twins for Deformable Objects from Egocentric Video
}

\author{Hyunjin Kim$^{1,2}$\hspace{4mm} Ri-Zhao Qiu$^{2}$ \hspace{4mm} Rocktim Jyoti Das$^{1}$ \hspace{4mm} Guangqi Jiang$^{2}$ \\[0.1cm] \hspace{4mm} Dinesh Manocha$^{1\dagger}$ \hspace{4mm} Xiaolong Wang$^{2\dagger}$ \\[0.3cm]
$^{1}$University of Maryland, College Park \hspace{4mm} $^{2}$UC San Diego%
\thanks{\textsuperscript{$\dagger$}Equal advising.}
}
\makeatletter
\providecommand{\captionof}[1]{\def\@captype{#1}\caption}
\makeatother

\IEEEaftertitletext{
    \begin{minipage}{\textwidth}
    \centering
    \includegraphics[width=1\textwidth]{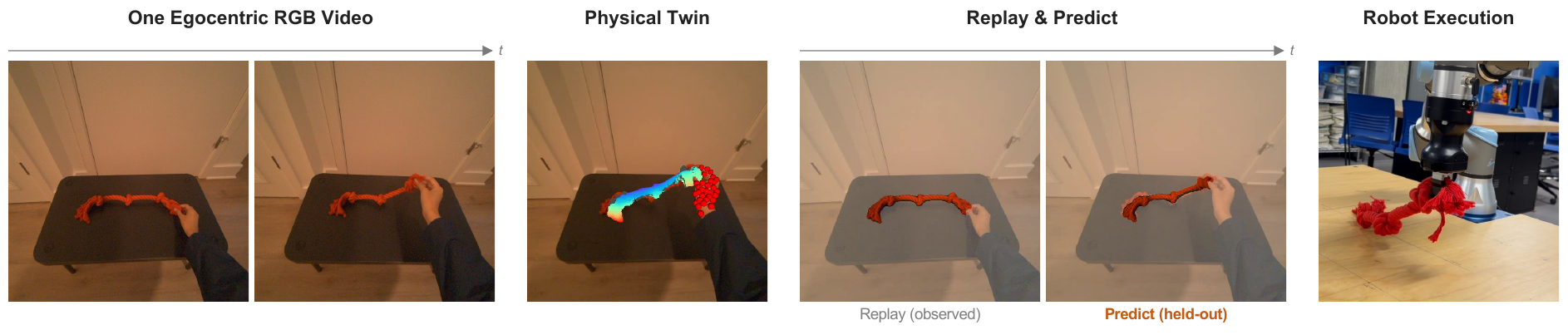}
    \vspace{-1.5em}
    \captionof{figure}{
\textbf{EgoPhys: Physical twins from a single egocentric video.}
From just one egocentric RGB video of a human interaction, EgoPhys identifies a
spring--mass physical twin of a deformable object, whose local stiffness is refined
by a frozen residual model learned from other interactions. The twin replays the
observed motion and predicts its held-out future, then plans actions that
a real robot executes without re-identifying the physical model.}
    \vspace{1em}
    \label{fig:teaser}
\end{minipage}
}

\begin{document}

\maketitle

\thispagestyle{empty}
\pagestyle{empty}

\begin{abstract}
Humans naturally understand object physics through everyday interactions, but predicting how deformable objects, such as soft objects and fabrics, respond to manipulation remains a major challenge for computer vision and robotics. Existing deformable physical twin methods typically rely on controlled multi-view or depth capture and costly per-object physical refinement. We present EgoPhys, a framework for constructing predictive deformable physical twins from a single egocentric RGB interaction. EgoPhys combines persistent correspondences with per-frame monocular 3D observations to fit a sequence-specific spring--mass model from moving-camera observations. Because a short interaction weakly constrains fine-grained local stiffness, EgoPhys amortizes this refinement across training interactions using a compact state-conditioned residual model that predicts bounded local corrections around the sequence-specific estimate and remains frozen at test time. Experiments show improved held-out future prediction over adapted deformable-twin baselines, with the amortized residual further improving prediction while reducing local-refinement cost by approximately 10--59$\times$ over dense per-spring optimization. The resulting twins also support deformable-object action selection, with representative physical robot manipulation providing a qualitative proof of concept. The project page is at \url{https://hjhyunjinkim.github.io/EgoPhys/}
\end{abstract}
\section{INTRODUCTION}
\label{sec:intro}

Predicting how deformable objects respond to interaction remains a significant problem in computer vision and robotics. For manipulation tasks such as folding cloth, handling bags, or repositioning soft objects, recovering an object's appearance or current shape is not enough: a robot must also predict how the object will deform under new actions. This is a \textit{system identification} problem: observations of an interaction must be used to recover a physical model whose parameters explain the observed motion and support prediction under future interactions~\cite{jatavallabhula2021gradsim,li2023pacnerf}. The goal is therefore not simply to reconstruct what happened, but to obtain a simulator that can predict what will happen next.

Existing deformable digital twin methods have made substantial progress toward this goal, but typically rely on controlled third-person capture, synchronized multi-view or RGB-D observations, known camera geometry, or other instrumented setups~\cite{li2023pacnerf,vasile2026gaussian,jiang2025phystwin,zhang2025real}. These requirements restrict physical data collection to carefully prepared environments. Egocentric cameras offer a complementary setting: they naturally observe hands, objects, and actions during everyday interaction without requiring external camera placement. However, this setting is substantially harder: the camera moves continuously, sensor depth is unavailable, visibility changes, and hand--object contact occludes precisely the regions where interaction forces are applied.

To address these challenges, we introduce \textbf{EgoPhys}, a framework for constructing deformable physical twins from a single egocentric RGB interaction. EgoPhys combines persistent image correspondences with per-frame monocular egocentric observations to obtain object-motion and manipulation constraints from a moving first-person camera. These observations instantiate a sequence-specific spring--mass simulator whose geometry and physical parameters are fitted to the observed interaction. EgoPhys further introduces egocentric-specific adaptations that make this inverse-physics process robust to moving viewpoints, incomplete geometry, and uncertain contact. In this way, EgoPhys extends explicit deformable real-to-sim from controlled capture to a substantially less constrained egocentric setting.

However, a short interaction may leave fine-grained physical parameters weakly constrained, making dense per-spring stiffness both difficult to estimate and expensive to optimize independently for every new interaction. Rather than densely optimizing an independent stiffness for every spring at test time, we amortize this fine-grained refinement through a compact shared residual model learned across multiple object interactions. Starting from the sequence-specific global stiffness, it predicts bounded local corrections from spring structure and the current simulated state. The model remains frozen at test time, supplementing rather than replacing object-specific system identification.

Experiments show that EgoPhys improves held-out future prediction over adapted deformable-twin baselines. The amortized residual refinement further improves independently identified twins on held-out interactions, including unseen physical instances, while outperforming dense per-spring refinement with $10$--$59\times$ lower test-time refinement cost. We additionally use the inferred simulators for deformable-object planning and illustrate open-loop physical robot manipulation.

We summarize our contributions as follows:
\begin{itemize}

\item \textbf{Deformable physical twins from egocentric RGB videos.} We construct predictive spring--mass simulators from a single egocentric RGB interaction, without synchronized depth sensing or fixed calibrated viewpoints.

\item \textbf{Amortized fine-grained physics refinement.}
We introduce a structured state-conditioned residual model that refines a newly identified sequence-specific simulator without dense per-spring test-time optimization. The model remains frozen at test time and predicts bounded relative spring-response corrections.

\item \textbf{Prediction and planning evaluation.}
We evaluate held-out future dynamics, compare against adapted deformable digital twin and per-scene refinement baselines, and use the inferred simulators for deformable-object planning and physical robot execution.

\end{itemize}

\section{RELATED WORK}
\label{sec:related}

\subsection{Deformable Physical Twins from Visual Observations}

Recent work combines visual reconstruction with physics-based simulation to recover simulatable models of deformable objects. Early approaches often assume pre-scanned geometry, clean point clouds, or controlled capture~\cite{li2023pacnerf}, while later methods couple learned 3D representations, including SDFs~\cite{qiao2022neuphysics}, NeRFs~\cite{chen2022virtual}, and Gaussian Splatting~\cite{jiang2024vr,zhang2024physdreamer,chopra2025physgs}, with differentiable simulation~\cite{Hu2020DiffTaichi,hu2019chainqueen}. Related methods use explicit spring--mass or elastodynamic models~\cite{zhong2024reconstruction,xie2024physgaussian,feng2024pie} to recover physical parameters alongside object geometry and motion. Together, these works show that visual observations can support predictive physical models, but typically assume observations acquired in structured sensing conditions.

Recent physical-twin systems move closer to real interaction data by combining sparse visual observations with explicit inverse physics. PhysTwin~\cite{jiang2025phystwin} and PhysWorld~\cite{yang2025physworld} reconstruct deformable dynamics from interaction videos and identify sequence-specific physical parameters, while soft-body real-to-sim methods~\cite{zhang2025real} use reconstructed twins for downstream robotic evaluation. EgoPhys advances this direction to a single egocentric RGB interaction. In this setting, the system must recover both the deformable object and the manipulation under continuous camera motion, missing sensor depth, changing visibility, and hand--object occlusion. EgoPhys reconstructs temporally coherent object and manipulator observations from this moving first-person view and uses them to construct and fit an explicit predictive spring--mass simulator.

\subsection{Shared Deformable Dynamics and Robot Simulation}

A complementary line of work learns reusable deformation dynamics across objects and interactions. Graph- and particle-based neural simulators~\cite{sanchez2020learning,pfaff2020learning,zhang2025particle} learn forward dynamics directly over meshes or particles, while AdaptiGraph~\cite{zhang2024adaptigraph} adapts learned dynamics using physical properties inferred from new interactions. GS-Dynamics~\cite{zhang2024dynamics} combines dynamic Gaussian tracking with learned graph dynamics for prediction and planning. Concurrently, MatPhys~\cite{yang2026matphys} predicts spring–mass parameters directly from single-view monocular video using a feed-forward material-aware model, without scene-specific inverse optimization at inference. EgoPhys instead handles moving egocentric observations and retains explicit per-sequence identification, using learning only to amortize fine-grained residual refinement.

Physical simulators are also increasingly used as forward models for robot evaluation and action selection. SIMPLER~\cite{li2024evaluating} studies simulation as a proxy for real-world manipulation performance, while SplatSim~\cite{qureshi2025splatsim}, GSWorld~\cite{jiang2025gsworld}, and large-scale simulation frameworks~\cite{chen2025robotwin,Mu_2025_CVPR} improve the realism and scalability of simulated robot environments. EgoPhys complements these efforts by constructing the forward model directly from egocentric human interaction rather than from a pre-existing simulator or an instrumented capture setup. We evaluate the resulting physical twins not only through held-out motion prediction, but also by using them for deformable-object action selection and executing representative planned trajectories on a physical robot.
\section{METHOD}
\label{sec:method}
\begin{figure*}[t]
    \centering
    \includegraphics[width=0.95\linewidth]{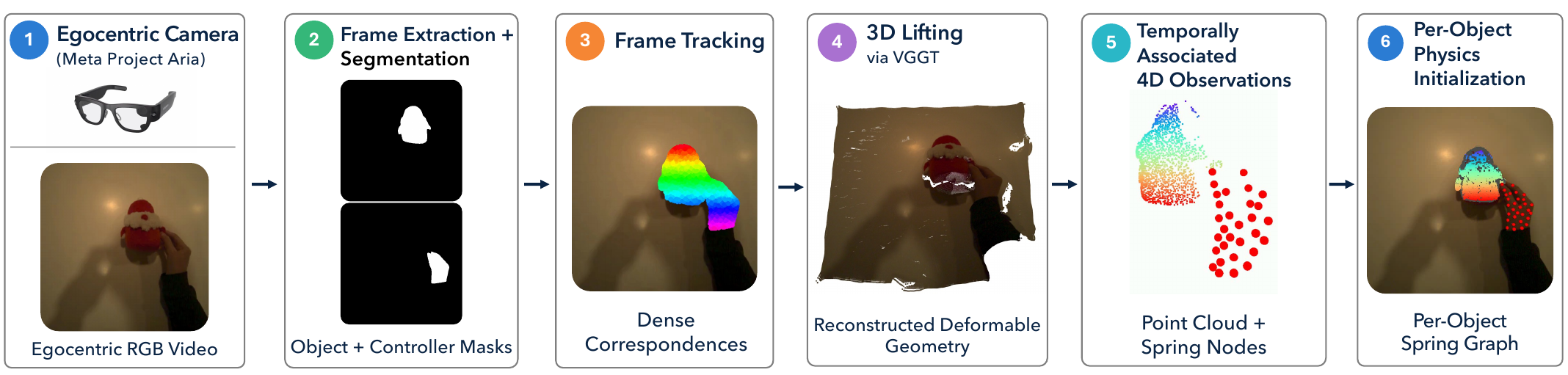}
    \caption{\textbf{Egocentric Motion Observation.} 
    From a single egocentric RGB video, EgoPhys segments and tracks the object and manipulator, combining persistent 2D correspondences with per-frame VGGT~\cite{wang2025vggt} monocular 3D predictions. These observations provide motion and manipulation constraints for sequence-specific spring--mass fitting.}
\label{fig:pipeline_data}
\end{figure*}

\begin{figure*}[t]
    \centering
    \includegraphics[width=\linewidth]{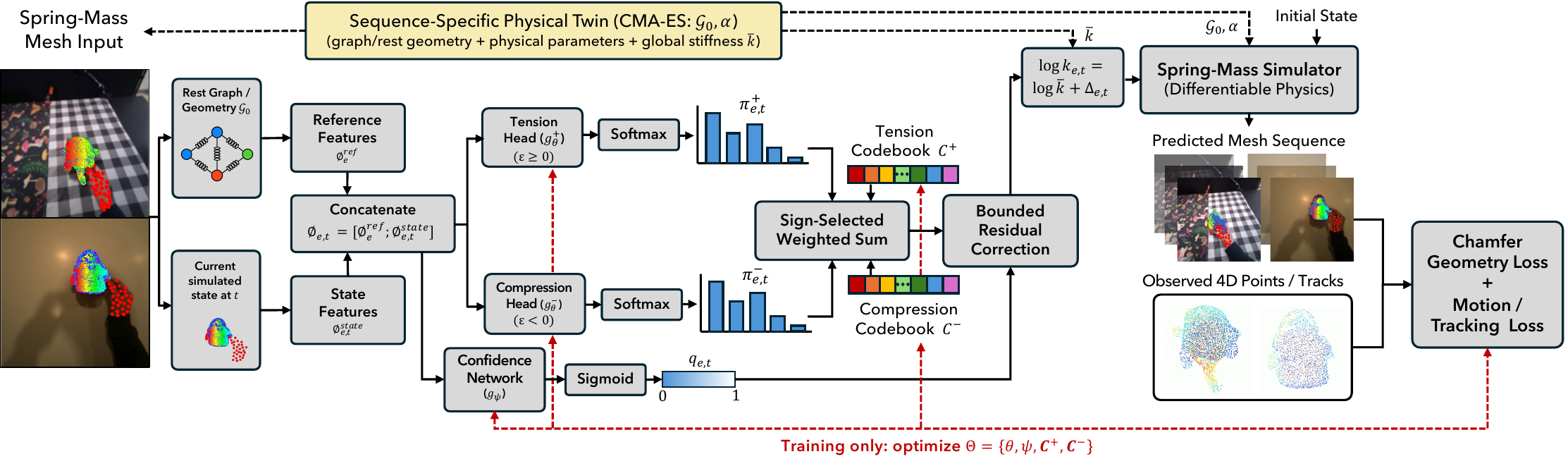}
    \caption{\textbf{Shared Residual Physics Model.}
    Starting from the sequence-specific stiffness estimate, we use  compact tension–compression codebooks to predict bounded local stiffness corrections from spring structure and simulated deformation state. The model is initialized on procedural spring graphs, fine-tuned on training interactions, and frozen at test time. Before each frame transition, the resulting effective stiffnesses are predicted and held fixed during integration.} 
    \label{fig:pipeline}
\end{figure*}

EgoPhys first combines persistent image
correspondences with monocular 3D predictions to recover object and manipulator
motion (Sec.~\ref{sec:data}). These observations are used to fit a
sequence-specific spring--mass model with adaptations for uncertain egocentric
geometry and contact (Sec.~\ref{sec:inverse_init}). However, since a short interaction
weakly constrains fine-grained local stiffness, we then refine this
sequence-specific estimate using a shared residual model learned across training
interactions and frozen at test time (Sec.~\ref{sec:codebook}).
Figs.~\ref{fig:pipeline_data} and~\ref{fig:pipeline} provide an overview.

\subsection{Egocentric Motion Observation}
\label{sec:data}

System identification for deformable objects requires observations of how the object moves and how it is manipulated over time. In controlled settings,
these quantities can be obtained from calibrated multi-view or RGB-D capture. EgoPhys instead recovers them from a single egocentric RGB sequence recorded
with a wearable camera (Meta Project Aria Gen 1)~\cite{engel2023project}.
Our observation pipeline builds on the segmentation, tracking, and controller-processing components of PhysTwin~\cite{jiang2025phystwin}, while adapting the reconstruction process to the egocentric monocular setting using persistent image correspondences and per-frame monocular 3D observations.

We first rectify the Aria RGB stream to a virtual pinhole view and obtain object
and manipulator masks with Grounded-SAM2~\cite{ren2024grounded}.
Within these regions,
CoTracker3~\cite{karaev2025cotracker3} provides persistent image-space point
identities and visibility across time, allowing the same object locations to be
associated across the moving egocentric view. For a visible tracked point with
image location $\mathbf u^t\in\mathbb R^2$ at frame $t$, we sample the
corresponding per-frame VGGT~\cite{wang2025vggt} point map $\mathbf W^t$ and
confidence map $\rho^t$ to obtain the 3D observation
$\mathbf p^t=\mathbf W^t[\mathbf u^t]\in\mathbb R^3$ and its confidence
$\rho^t[\mathbf u^t]$. We retain observations with
$\rho^t[\mathbf u^t]\geq0.5$ and predicted depth
$0.2<(\mathbf p^t)_z<1.5$, and reject tracks that leave the semantic mask,
form isolated 3D outliers, or move inconsistently with neighboring tracks.

Together, these components provide temporally associated 4D observations without requiring global registration. VGGT is applied independently at each timestamp, so its observations are camera-referenced, while CoTracker3 provides persistent point identity. Since object and manipulator observations within each frame share the same camera coordinate system, frame-wise relative geometry and hand--object relationships are preserved. We use these correspondence-indexed observations as constraints for physical identification rather than as a globally registered
scene reconstruction. Following PhysTwin~\cite{jiang2025phystwin}, we separate object and hand regions and select control points by farthest-point sampling. When severe occlusion leaves the object's reference geometry incomplete, we use TRELLIS~\cite{xiang2025structured} to complete the first-frame shape and register it to the observed geometry, while the subsequent deformation remains constrained by the tracked observations. Finally, the lowest first-frame object points provide an estimate of the support surface. This gives EgoPhys the reference geometry, object motion, and manipulation trajectory needed to instantiate the sequence-specific physical twin.

\subsection{Egocentric Physical Twin Identification}
\label{sec:inverse_init}

We next convert the correspondence-indexed observations from
Sec.~\ref{sec:data} into a predictive physical model. 
We build EgoPhys on the explicit spring--mass simulator and trajectory-matching formulation of PhysTwin~\cite{jiang2025phystwin}. The underlying inverse-physics problem remains the same, but the egocentric observations in Sec.~\ref{sec:data} change how the physical system can be reliably instantiated. In particular, monocular reconstruction introduces uncertainty in hand--object alignment, support geometry, and the structure of thin deformable objects. We therefore adapt the simulator construction to these failure modes before performing sequence-specific identification.

First, egocentric monocular reconstruction errors can slightly separate the observed hand and object, causing radius-based controller attachment to fail to detect contact. We retain radius-based neighbors when available, but otherwise attach the nearest object particles within a maximum distance and initialize controller-spring rest lengths from their observed separation. This preserves manipulation coupling under small hand--object misalignment.

Second, egocentric monocular observations do not provide a known support height, so using PhysTwin's fixed zero-height ground plane can place the object incorrectly relative to its support surface. We instead estimate the support height from the first-frame object observations, $z_g=\min_i[\mathbf x_{i,0}]_z$, and use $z=z_g$ for ground contact while preserving the observed vertical coordinates. This allows lifted regions to remain above the support rather than being projected onto a fixed ground plane.

Third, thin objects such as towels and bags are particularly sensitive to the partial geometry recovered from egocentric monocular observations. A graph with only short-range springs can fold or crumple unrealistically, while tracking errors can cause excessive tensile deformation. We add lower-stiffness long-range bending connections within $r_b=2.2r_o$, where $r_o$ is the local object-spring radius, and set their stiffness to $15\%$ of the local spring stiffness. We also limit excessive tensile strain while retaining compression, stabilizing thin-object deformation under incomplete geometry.

With these egocentric-specific constructions in place, object particles form
the nodes of a spring graph
$\mathcal G=(\mathcal V,\mathcal E)$, initialized from the recovered reference
geometry. The force on node $i$ is
\begin{equation}
\begin{aligned}
\mathbf F_i
={}&
\sum_{(i,j)\in\mathcal E}
\Bigg[
k_{ij}
\left(
\frac{\|\mathbf x_j-\mathbf x_i\|}{l_{ij}}-1
\right)
\hat{\mathbf d}_{ij}
\\[-2pt]
&\qquad
-\gamma
\left(
(\mathbf v_i-\mathbf v_j)^\top
\hat{\mathbf d}_{ij}
\right)
\hat{\mathbf d}_{ij}
\Bigg]
+\mathbf F_i^{\mathrm{ext}},
\end{aligned}
\label{eq:force}
\end{equation}
where $\mathbf x_i$ and $\mathbf v_i$ are the position and velocity of particle
$i$, $l_{ij}$ is the rest length of spring $(i,j)$, and
$\hat{\mathbf d}_{ij}$ is its current unit direction. The stiffness $k_{ij}$
controls the elastic response and $\gamma$ controls damping along the spring.
$\mathbf F_i^{\mathrm{ext}}$ collects gravity, collisions, support contact, and
forces transmitted through the observed manipulation. During sequence-specific
identification, object springs share a global stiffness
$k_{ij}=\bar{k}$.

The recovered manipulator trajectory provides the control input $a_t$, and we
roll the simulator forward using semi-implicit Euler integration,
$
    \hat{\mathbf X}_{t+1}
    =
    f_{\alpha,\mathcal G_0}
    (\hat{\mathbf X}_t,a_t),
    \label{eq:sim_rollout}
$
where $\hat{\mathbf X}_t$ denotes the simulated object state and
$\mathcal G_0$ contains the sequence-specific connectivity and rest geometry.
The parameter vector $\alpha$ collects the physical and contact quantities
identified from the current interaction, including global stiffness $\bar{k}$,
damping, collision parameters, strain limits, and ground-contact parameters. The particle masses and gravity are fixed.

Our goal is to find $\mathcal{G}_0$ and $\alpha$ whose simulated rollout best matches the observed interaction. We identify them by minimizing a trajectory-matching objective over the observed frames:
\begin{equation}
\mathcal L_{\mathrm{seq}}
=
\sum_t
\left[
\lambda_{\mathrm{geo}}\mathcal C_{\mathrm{geo}}
+
\lambda_{\mathrm{mot}}\mathcal C_{\mathrm{mot}}
+
\lambda_{\mathrm{acc}}\mathcal C_{\mathrm{acc}}
\right],
\label{eq:coarse_loss}
\end{equation}
where $\mathcal C_{\mathrm{geo}}$ is a one-sided squared Chamfer loss from the visible object observations to the simulated surface, and $\mathcal C_{\mathrm{mot}}$ is a Smooth-$L_1$ loss over valid tracked correspondences. $\mathcal C_{\mathrm{acc}}$ compares each particle's current velocity change, $\mathbf v_{i,t+1}-\mathbf v_{i,t}$, with its previous change, $\mathbf v_{i,t}-\mathbf v_{i,t-1}$, using Smooth-$L_1$. It therefore penalizes abrupt changes in acceleration (temporal jitter), rather than the object's motion itself. We use $(\lambda_{\mathrm{geo}},\lambda_{\mathrm{mot}}, \lambda_{\mathrm{acc}})=(1,1,0.01)$ and jointly search over graph-construction, physical, and contact parameters using the covariance matrix adaptation evolution strategy (CMA-ES)~\cite{hansen2016cma}.

This produces a sequence-specific physical twin with an identified spring graph, rest geometry, global stiffness $\bar{k}$, damping, collision, strain-limit, and contact parameters. Appearance reconstruction follows PhysTwin~\cite{jiang2025phystwin}. Importantly, this stage already provides a predictive object-specific model. The remaining difficulty is finer-grained: the same short monocular interaction provides much weaker evidence for an independent stiffness at every spring. Rather than optimizing these degrees of freedom separately for each sequence, we amortize this refinement across training interactions from multiple physical object instances.

\subsection{Shared Residual Physics Model}
\label{sec:codebook}
The sequence-specific identification in Sec.~\ref{sec:inverse_init} recovers a predictive physical model for the current interaction, including its global stiffness $\bar{k}$. However, a short and partially observed egocentric interaction provides much weaker evidence for how stiffness should vary locally across individual springs. Estimating these fine-grained parameters independently for every new sequence therefore introduces many weakly constrained degrees of freedom and requires additional test-time optimization. EgoPhys instead shares only this fine-grained refinement across interactions. We retain the sequence-specific global estimate as an object-specific anchor and learn a compact, state-conditioned residual model that predicts bounded local deviations around it. The shared model is trained once and remains frozen at test time, thereby supplementing rather than replacing sequence-specific physical identification.

For spring $e=(i,j)$ at frame $t$, we write its effective stiffness as
\begin{equation}
    \log k_{e,t}
    =
    \log\bar{k}+\Delta_{e,t},
    \qquad
    \Delta_{e,t}
    =
    \mathcal P_{\Theta}(\boldsymbol{\phi}_{e,t}),
    \label{eq:anchored_codebook}
\end{equation}
where $k_{e,t}$ is the stiffness used from frame $t$ to $t+1$, $\Delta_{e,t}$ is a residual correction in log-stiffness space, $\boldsymbol{\phi}_{e,t}=[\boldsymbol{\phi}^{\mathrm{ref}}_{e};\boldsymbol{\phi}^{\mathrm{state}}_{e,t}]$ concatenates the reference and state features, and $\mathcal P_{\Theta}$ is the shared predictor. Its parameters are $\Theta=\{\theta,\psi,\mathbf C^+,\mathbf C^-\}$, defined below. Thus, $\Delta_{e,t}=0$ recovers the sequence-specific estimate $\bar{k}$, while positive and negative residuals locally increase or decrease stiffness.

The reference features $\boldsymbol{\phi}^{\mathrm{ref}}_e$ encode time-invariant properties of the spring in its reference configuration, including rest geometry, endpoint connectivity, spring type, and normalized local shape. The state features $\boldsymbol{\phi}^{\mathrm{state}}_{e,t}$ encode the spring's current simulated configuration and deformation, including orientation relative to gravity, height above the support plane, axial strain $\epsilon_{e,t}$, and strain rate $\dot{\epsilon}_{e,t}$. The axial strain measures relative length change from rest, while strain rate measures its rate of change:
\begin{equation}
    \epsilon_{e,t}
    =
    \frac{\|\mathbf x_{j,t}-\mathbf x_{i,t}\|}{l_e}-1,
    \qquad
    \dot{\epsilon}_{e,t}
    =
    \frac{
    (\mathbf v_{j,t}-\mathbf v_{i,t})^\top
    \hat{\mathbf d}_{e,t}
    }{l_e}.
    \label{eq:strain_features}
\end{equation}
Here, $l_e$ is the rest length, $\mathbf{x}_{i,t}$ and $\mathbf{v}_{i,t}$ are the simulated position and velocity of endpoint $i$, and $\hat{\mathbf d}_{e,t}$ is the current spring direction (positive strain: tension; negative: compression). Since the prediction depends on simulated state, $k_{e,t}$ is an effective, not intrinsic, stiffness. Object springs use both reference and state features, while controller springs use reference features together with a controller indicator.

Rather than directly regressing an arbitrary residual for each spring, we restrict the correction to mixtures of shared responses. We use two compact codebooks, $\mathbf C^+=[c^+_1,\ldots,c^+_K]$ for tension and $\mathbf C^-=[c^-_1,\ldots,c^-_K]$ for compression, whose entries are candidate log-stiffness corrections. A shared assignment network $g_\theta$ has separate tension and compression heads and produces mixture weights
\begin{equation}
    \boldsymbol{\pi}^{\pm}_{e,t}
    =
    \operatorname{softmax}\!\left(
    g_{\theta}^{\pm}(\boldsymbol{\phi}_{e,t})/\tau
    \right),
\end{equation}
where $\tau$ is the softmax temperature. The sign of the current strain selects the corresponding codebook,
\begin{equation}
    \Delta^{\mathrm{raw}}_{e,t}
    =
    \begin{cases}
        (\boldsymbol{\pi}^{+}_{e,t})^\top\mathbf C^{+},
        & \epsilon_{e,t}\geq0,\\
        (\boldsymbol{\pi}^{-}_{e,t})^\top\mathbf C^{-},
        & \epsilon_{e,t}<0.
    \end{cases}
    \label{eq:codebook_assign}
\end{equation}

Additionally, a confidence network $g_\psi$ controls how strongly the shared correction modifies the sequence-specific estimate,
\begin{equation}
    q_{e,t}
    =
    \sigma(g_\psi(\boldsymbol{\phi}_{e,t})),
    \qquad
    \Delta_{e,t}
    =
    q_{e,t}\Delta_{\max}
    \tanh\!\left(
    \frac{\Delta^{\mathrm{raw}}_{e,t}}{\Delta_{\max}}
    \right),
\end{equation}
where $\Delta_{\max}=0.7$ bounds the magnitude of the residual correction. The bounded correction $\Delta_{e,t}$ is then added to the sequence-specific log-stiffness through Eq.~\ref{eq:anchored_codebook}, giving $k_{e,t}=\bar{k}\exp(\Delta_{e,t})$. This effective stiffness is used by the simulator for the transition from frame $t$ to $t+1$. 

We learn the residual model in two stages. First, we train the
assignment network and codebooks on procedural spring graphs using MSE between the predicted and known log-stiffness residuals, plus the regularizers below. Second, we introduce the confidence network and fine-tune the complete model through differentiable rollouts of real training interactions. Holding each interaction's identified global stiffness $\bar{k}^{(o)}$ fixed, we minimize
\begin{equation}
\begin{aligned}
\mathcal L_{\mathrm{res}}
&=\mathcal L_{\mathrm{roll}}
+\lambda_\Delta\mathcal R_\Delta
+\lambda_{\mathrm{use}}\mathcal R_{\mathrm{use}}
+\lambda_{\mathrm{ent}}\mathcal R_{\mathrm{ent}},\\
\mathcal L_{\mathrm{roll}}
&=\sum_{o,t}\left[
\mathcal C_{\mathrm{geo}}
(\hat{\mathbf X}^{(o)}_t,\mathbf X^{(o)}_t)
+\mathcal C_{\mathrm{mot}}
(\hat{\mathbf X}^{(o)}_t,\mathbf X^{(o)}_t)\right].
\end{aligned}
\label{eq:codebook_training}
\end{equation}
Let $\mathbf C^\pm$ be the $K$-entry codebooks, $\pi^s_{mk}$ the assignment of sample $m$ to entry $k$, and $\bar{\pi}^s_k=M^{-1}\sum_m\pi^s_{mk}$. We use
\begin{equation}
\begin{aligned}
\mathcal R_\Delta&=\frac{\|\mathbf C^+\|_2^2+\|\mathbf C^-\|_2^2}{2K},\\
\mathcal R_{\mathrm{use}}&=\frac{1}{K}\sum_{s,k}
(\bar{\pi}^s_k-K^{-1})^2,\\
\mathcal R_{\mathrm{ent}}&=-\frac{1}{M}\sum_{s,m,k}
\pi^s_{mk}\log(\pi^s_{mk}+\epsilon).
\end{aligned}
\end{equation}
These regularizers keep offsets near zero, balance codebook usage, and encourage concentrated assignments. Their weights are $(10^{-5},10^{-3},10^{-4})$ in the first stage and $(10^{-4},10^{-3},10^{-4})$ in the second.

At test time, we identify the sequence-specific physical twin once and fix its global stiffness $\bar{k}$. The residual model remains frozen and predicts $\Delta_{e,t}$ before each transition, giving $k_{e,t}=\bar{k}\exp(\Delta_{e,t})$ without per-spring test-time optimization. Thus, EgoPhys preserves object-specific physical identification while sharing only the fine-grained, state-conditioned refinement across interactions.
\section{EXPERIMENTS}
\label{sec:experiments}
Our experiments follow EgoPhys's two-stage design. We first isolate sequence-specific identification, evaluating whether the egocentric adaptations recover a more predictive physical twin from the same visual observations. We then isolate the shared residual model, evaluating whether fine-grained refinement improves held-out prediction, how it compares with direct residual
prediction, and whether dense per-sequence optimization is an effective alternative. Finally, we evaluate the resulting twins as forward models for deformable-object planning and physical execution.

\subsection{Experimental Setup}
\begin{figure*}[t]
    \centering
    \includegraphics[width=\linewidth]{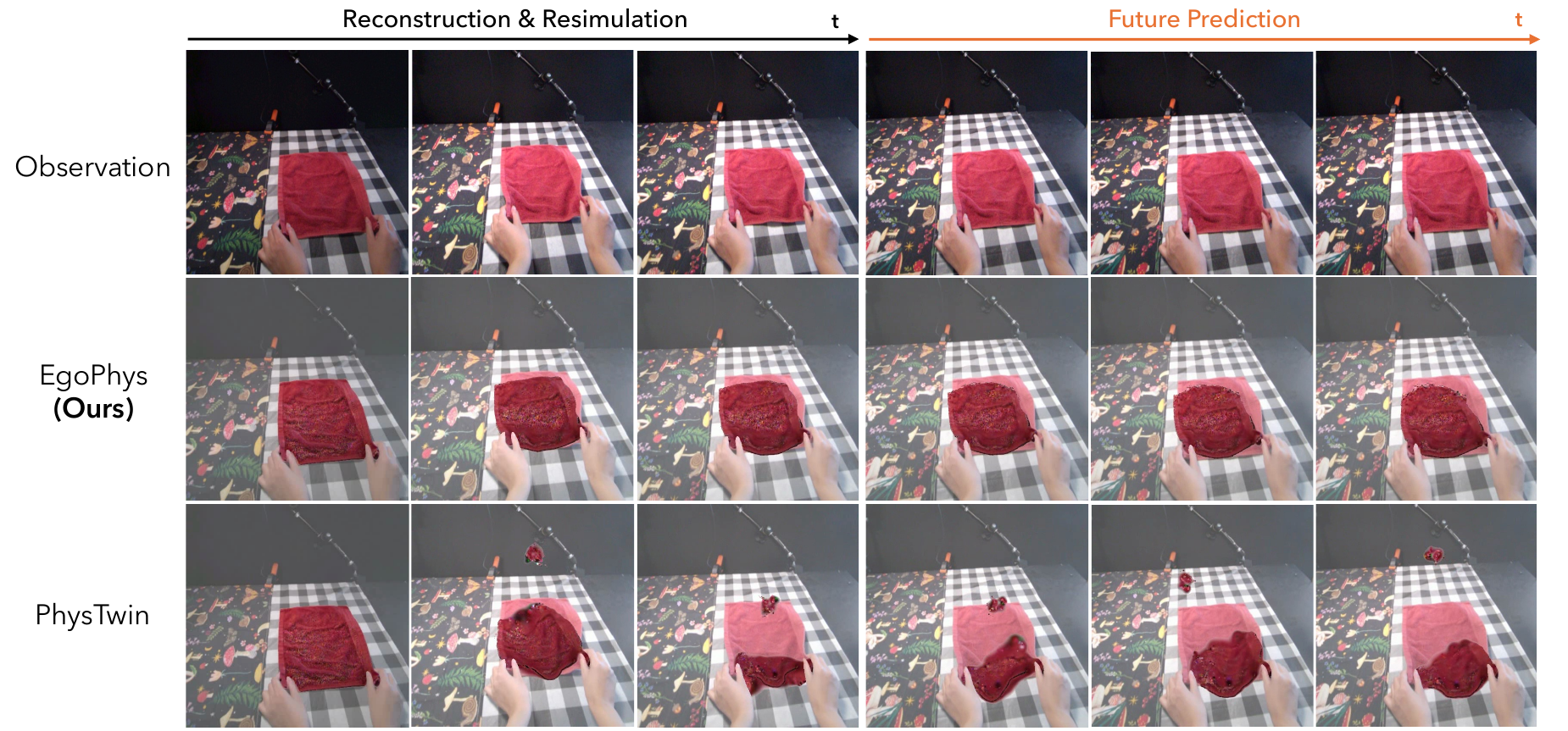}
    \caption{\textbf{Egocentric physical-twin resimulation and future prediction.} On a representative towel-pulling interaction, the inferred EgoPhys simulator more closely follows the observed deformation during resimulation and future prediction.}
    \label{fig:qualitative}
\end{figure*}
\begin{table*}[t]
\caption{\textbf{Sequence-specific physical identification from Egocentric RGB.} All methods receive the same recovered egocentric observations and are fitted using the observed 70\% of each interaction. The shared residual model is disabled for EgoPhys in this comparison.}
\centering
\setlength{\tabcolsep}{3.2pt}
\renewcommand{\arraystretch}{0.95}

\begin{tabular}{@{}l@{\hspace{4pt}}rrrrrr@{\hspace{8pt}}rrrrrr@{}}
\toprule
& \multicolumn{6}{c}{Resimulation (70\%)}
& \multicolumn{6}{c}{Future Prediction (30\%)} \\
\cmidrule(lr){2-7} \cmidrule(l){8-13}
Method
& CD $\downarrow$ & TE $\downarrow$ & IoU\% $\uparrow$
& PSNR $\uparrow$ & SSIM $\uparrow$ & LPIPS $\downarrow$
& CD $\downarrow$ & TE $\downarrow$ & IoU\% $\uparrow$
& PSNR $\uparrow$ & SSIM $\uparrow$ & LPIPS $\downarrow$ \\
\midrule
Spring-Gaus~\cite{zhong2024reconstruction}
& 7.250 & 0.844 & 47.5 & 22.218 & 0.943 & 0.113
& 32.924 & 2.064 & 26.6 & 20.033 & 0.936 & 0.142 \\
PhysTwin~\cite{jiang2025phystwin}
& 0.024 & 0.035 & 71.4 & 24.343 & \textbf{0.946} & 0.061
& 0.034 & 0.064 & 52.4 & 21.907 & \textbf{0.939} & 0.092 \\
\midrule
EgoPhys \textbf{(Ours)}
& \textbf{0.015} & \textbf{0.025} & \textbf{77.8}
& \textbf{24.849} & \textbf{0.946} & \textbf{0.052}
& \textbf{0.028} & \textbf{0.052} & \textbf{58.8}
& \textbf{22.019} & 0.938 & \textbf{0.087} \\
\bottomrule
\end{tabular}
\label{tab:main}
\end{table*}
\paragraph{Dataset and Splits}
We collect and evaluate on 19 egocentric deformable-object interactions, each containing 210 RGB frames (approximately 7\,s), for 3,990 frames in total. Each interaction is split chronologically into 147 observed frames (70\%) for sequence-specific system identification and 63 future frames (30\%) for rollout evaluation. Thus, each physical interaction provides a dense temporal trajectory rather than a single supervision signal.

The shared residual model is initialized on procedural spring graphs and fine-tuned on 1,176 observed frames from eight training interactions (1,168 ordered state transitions), with the remaining 504 frames used only for checkpoint selection. We then freeze the model and evaluate on 11 held-out interactions: four with a previously seen physical instance under a new action and seven with previously unseen instances. For these interactions, 1,617 observed frames are used only for sequence-specific identification and 693 future frames only for evaluation. Three of the seven unseen-instance interactions form a stricter towel-family subset, with no towel interaction used for residual training or selection. Thus, for each held-out interaction, the base physical parameters are identified using only that interaction's observed frames, while the frozen residual model only refines this estimate using patterns learned from the training interactions.

\paragraph{Baselines and Metrics}
To our knowledge, no prior method constructs deformable physical twins from a single egocentric RGB interaction. We therefore compare against PhysTwin~\cite{jiang2025phystwin}, the closest explicit deformable physical-twin method, and Spring-Gaus~\cite{zhong2024reconstruction}, adapting their input interfaces to consume the same recovered observations, masks, correspondences, trajectories, and temporal split as EgoPhys. We otherwise retain their underlying physical modeling procedures. This isolates how well each method fits a physical model from identical egocentric evidence. We exclude PhysWorld~\cite{yang2025physworld}, which lacks a public implementation.

We report Chamfer distance (CD) for geometric agreement, tracking error (TE) for motion accuracy over correspondences, and silhouette IoU for image-space overlap, along with PSNR, SSIM, and LPIPS for rendered appearance. Human-occluded pixels are excluded from image-space evaluation. We compute metrics per frame, average within each interaction, then macro-average across interactions to weight interactions equally. We emphasize the held-out future 30\%, since reproducing fitting frames alone does not establish predictive dynamics.
\begin{figure*}[thpb]
    \centering
    \includegraphics[width=0.95\linewidth]{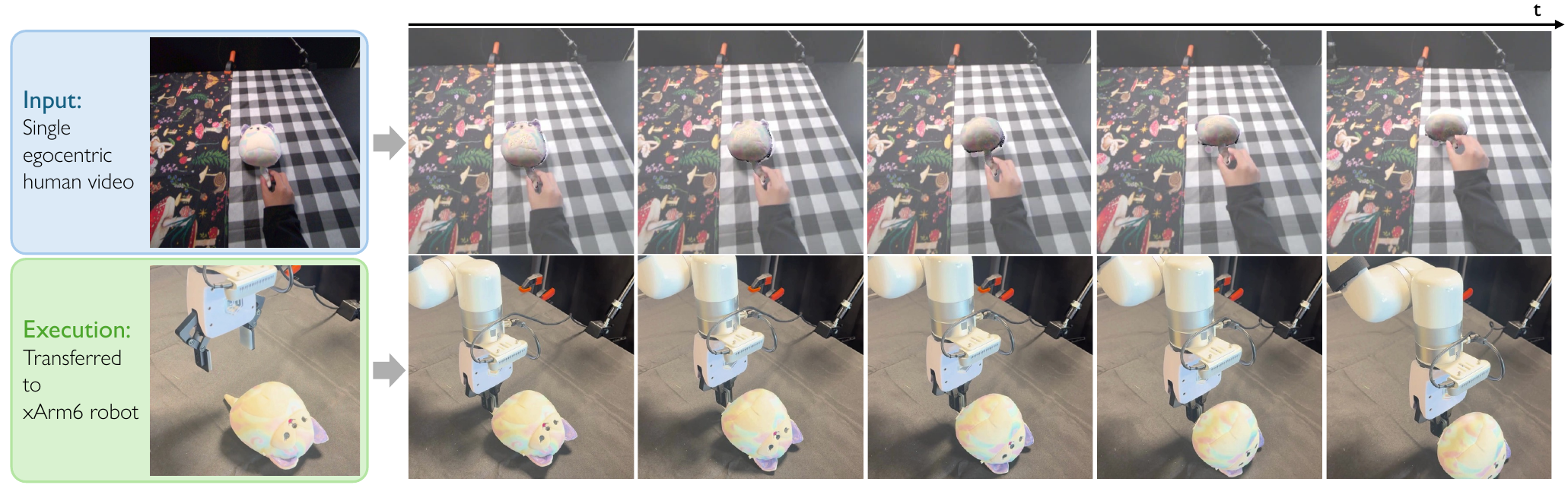}
    \caption{\textbf{Physical execution of EgoPhys-planned interactions.}
    Representative trajectories optimized with the inferred EgoPhys
    physical twins are executed on an xArm6 robot without re-identifying
    the physical model. The resulting interactions move the deformable objects
    toward their target configurations.}
    \label{fig:robot}
\end{figure*}
\subsection{Do Egocentric-Specific Adaptations Improve Physical Twin Prediction?}
We first evaluate whether EgoPhys's egocentric-specific design choices yield more accurate predictive physical twins. Building such a model from first-person RGB is challenging: the camera moves with the user, depth is unavailable, and hand–object contact is often ambiguous. EgoPhys addresses this by combining persistent image correspondences with per-frame monocular observations to construct a spring–mass simulator with robust hand–object coupling, scene-derived support contact, and thin-object stabilization, producing a usable physical system directly from egocentric interaction, without calibrated RGB-D.

All methods receive identical recovered observations, correspondences, trajectories, and temporal splits. As Tab.~\ref{tab:main} shows, EgoPhys achieves better overall resimulation and held-out future prediction than the adapted PhysTwin and Spring-Gaus baselines. The future prediction gain matters most, since it shows that the recovered physical model remains predictive beyond the fitting frames. Fig.~\ref{fig:qualitative} illustrates this on a towel interaction: EgoPhys better preserves deformation during resimulation and continues to track object motion during prediction. The accompanying video includes more qualitative comparisons with Spring-Gaus~\cite{zhong2024reconstruction} and PhysTwin~\cite{jiang2025phystwin}.

\subsection{Does the Shared Residual Improve Held-Out Prediction?}

We next isolate the frozen shared residual model on the held-out interactions described above. For each interaction, we take the sequence-specific physical twin independently identified in Sec.~\ref{sec:inverse_init} and evaluate it with and without residual refinement. Neither variant performs dense per-spring optimization. As a cross-domain baseline, we evaluate zero-shot MatPhys~\cite{yang2026matphys}, a feed-forward model trained on fixed-camera single-view videos. In contrast, our moving egocentric camera requires 4D observations under changing viewpoints and visibility. We use its checkpoint without egocentric retraining or per-sequence optimization and evaluate the predicted parameters in our spring--mass rollout.
\begin{table}[t]
\caption{\textbf{Shared residual refinement on held-out interactions.} EgoPhys adds a frozen shared residual to the sequence-specific twin. MatPhys is evaluated zero-shot without egocentric retraining or per-sequence optimization. The towel subset contains 3 of the 7 unseen interactions and is excluded from residual training and checkpoint selection.
}
\centering
\fontsize{7.1}{8.3}\selectfont
\setlength{\tabcolsep}{3.0pt}
\renewcommand{\arraystretch}{0.94}
\begin{tabular}{@{}llrrrr@{}}
\toprule
Evaluation group & Method & $N$
& CD $\downarrow$ & TE $\downarrow$ & IoU $\uparrow$ \\
\midrule
\multirow{3}{*}{Seen inst., new action}
& MatPhys~\cite{yang2026matphys} & 4 & 0.1789 & 0.1841 & 33.3 \\
& Sequence-specific twin & 4 & 0.0275 & 0.0599 & 52.6 \\
& + shared residual & 4 & \textbf{0.0267} & \textbf{0.0543}
& \textbf{53.8} \\
\midrule
\multirow{3}{*}{Unseen instance}
& MatPhys~\cite{yang2026matphys} & 7 & 0.1416 & 0.1566 & 34.9 \\
& Sequence-specific twin & 7 & 0.0414 & 0.0752 & 53.9 \\
& + shared residual & 7 & \textbf{0.0349} & \textbf{0.0666}
& \textbf{57.1} \\
\midrule
\multirow{3}{*}{\shortstack[l]{Towel subset\\(Subset of unseen)}}
& MatPhys~\cite{yang2026matphys} & 3 & 0.0723 & \textbf{0.0905} & 55.6 \\
& Sequence-specific twin & 3 & 0.0666 & 0.1110 & 52.7 \\
& + shared residual & 3 & \textbf{0.0544} & 0.0936
& \textbf{56.7} \\
\midrule
\multirow{3}{*}{All held-out}
& MatPhys~\cite{yang2026matphys} & 11 & 0.1552 & 0.1666 & 34.3 \\
& Sequence-specific twin & 11 & 0.0364 & 0.0696 & 53.5 \\
& + shared residual & 11 & \textbf{0.0319} & \textbf{0.0621}
& \textbf{55.9} \\
\bottomrule
\end{tabular}
\label{tab:heldout_transfer}
\end{table}

Tab.~\ref{tab:heldout_transfer} shows that EgoPhys improves mean future prediction over the sequence-specific twin across all 11 held-out interactions, comprising four interactions with a seen instance under a new action and seven with unseen instances. The mean improvement also holds on the three-interaction towel subset within the unseen group, for which no towel interaction is used during residual training or checkpoint selection. By contrast, the MatPhys~\cite{yang2026matphys} predictor transfers poorly zero-shot to our deformable plush objects, although it remains competitive on towels.

\paragraph{Does shared residual refinement improve over direct per-spring prediction?}
We next test whether the held-out improvement simply follows from adding a learned per-spring predictor. All variants start from the same sequence-specific physical model. The direct MLP residual replaces the codebooks and confidence gate with a one-hidden-layer, 64-unit ReLU MLP using the same reference and state-dependent spring features and directly predicts tension- or compression-specific per-spring log stiffness. We train it from scratch on the same real interactions with EgoPhys's rollout losses.
\begin{table}[t]
\caption{\textbf{Ablation of the residual model.}
All variants start from the same sequence-specific physical model. Direct MLP
regression predicts per-spring residuals without the structured codebook
representation.}
\centering
\setlength{\tabcolsep}{4.2pt}
\renewcommand{\arraystretch}{0.95}
\begin{tabular}{@{}lrr@{}}
\toprule
Residual model
& Test CD $\downarrow$
& Test TE $\downarrow$ \\
\midrule
No residual refinement
& 0.0364 & 0.0696 \\
Direct MLP residual
& 0.0516 & 0.0889 \\
EgoPhys (shared residual)
& \textbf{0.0319} & \textbf{0.0621} \\
\bottomrule
\end{tabular}
\label{tab:codebook_ablation}
\end{table}
\begin{table}[t]
\caption{\textbf{Shared residual model versus PhysTwin-style dense
refinement.} Both start from the same sequence-specific physical twin.
EgoPhys applies a frozen shared residual model, while dense refinement
optimizes per-spring stiffnesses independently for each test interaction.
EgoPhys achieves better prediction with substantially lower test-time cost.}
\centering
\setlength{\tabcolsep}{2.8pt}
\renewcommand{\arraystretch}{0.95}
\begin{tabular}{@{}lcrrrrr@{}}
\toprule
& & \multicolumn{2}{c}{All held-out}
& \multicolumn{2}{c}{Towel family} & \\
\cmidrule(lr){3-4}
\cmidrule(lr){5-6}
Method
& Steps
& CD $\downarrow$
& TE $\downarrow$
& CD $\downarrow$
& TE $\downarrow$
& Time $\downarrow$ \\
\midrule
Shared residual
& 0
& \textbf{0.0319}
& \textbf{0.0621}
& \textbf{0.0544}
& \textbf{0.0936}
& \textbf{1.36 min} \\
\midrule
Dense refinement
& 25
& 0.0408
& 0.0761
& 0.0877
& 0.1437
& 14.26 min \\
Dense refinement
& 50
& 0.0404
& 0.0761
& 0.0864
& 0.1451
& 24.11 min \\
Dense refinement
& 100
& 0.0406
& 0.0798
& 0.0870
& 0.1595
& 43.99 min \\
Dense refinement
& 200
& 0.0405
& 0.0782
& 0.0863
& 0.1524
& 80.43 min \\
\bottomrule
\end{tabular}
\label{tab:calibration_budget}
\end{table}

Tab.~\ref{tab:codebook_ablation} shows that unconstrained residual prediction can hurt an already strong sequence-specific estimate. A direct MLP can assign independent corrections from noisy, partial observations, shifting local stiffness away from the identified physical model and degrading future prediction. EgoPhys instead restricts refinement to shared codebook responses and gates their magnitude, improving geometry and tracked motion while preserving the sequence-specific estimate.

\paragraph{Can shared refinement replace dense per-sequence optimization?}
PhysTwin uses a hierarchical coarse-to-dense identification strategy: after estimating topology and homogeneous physical parameters, it further optimizes dense per-spring stiffnesses for each sequence~\cite{jiang2025phystwin}. EgoPhys retains the sequence-specific coarse identification, but replaces this second-stage dense optimization with the frozen residual model. We therefore compare the two refinement strategies from the same sequence-specific physical twin while varying the dense optimization budget.

Tab.~\ref{tab:calibration_budget} shows that the dense per-spring refinement used by PhysTwin does not recover the predictive improvement of the shared residual model in our egocentric setting. Dense refinement introduces many local degrees of freedom that must be estimated from the same short, partial monocular interaction. In contrast, EgoPhys constrains these corrections using patterns learned from procedural graphs and multiple training interactions. Increasing dense refinement from 25 to 200 steps produces little improvement, indicating that the gap is not simply due to insufficient optimization budget. The shared residual also reduces local-refinement time from 14.26--80.43 minutes to 1.36 minutes, a reduction of approximately $10$--$59\times$ while also achieving better held-out prediction.

\subsection{Can EgoPhys Support Planning and Physical Execution?}

We finally ask whether the improved predictive twins provide useful forward models for action selection. We use both EgoPhys and adapted PhysTwin for model-predictive path integral (MPPI) planning~\cite{williams2015model} on 11 held-out interactions. For each interaction, we perform three planning runs with different random seeds, yielding 33 runs per method. Actions are 3D translations of one manipulation handle for single-hand interactions and two handles for double-hand interactions. We use 12 control keyframes, 24 samples, and six MPPI iterations, giving 144 candidate rollouts per run. Both methods use identical targets, action constraints, random seeds, planner parameters, and rollout budgets.

To avoid rewarding self-consistency, each method first optimizes actions using its own forward model. EgoPhys uses its inferred topology and contact formulation, while PhysTwin uses its original graph and contact model. We then replay both optimized action sequences unchanged in the same evaluator. For each sequence, the evaluator starts from the sequence-specific CMA-ES anchor and refines dense per-spring stiffnesses and ground/object contact parameters for 100 Adam iterations on the 147 observed frames. These optimized, not measured, parameters are then frozen for evaluation. We report final geometric error (CD), tracked-point error (TE), and $|\Delta_{\mathrm{self}}|$, which is the gap between each planner's predicted and evaluated CD. Success is defined as at least a 50\% reduction in final CD.

\begin{table}[t]
\caption{\textbf{Forward-model evaluation for deformable-object planning.} Actions planned with each physical twin are replayed unchanged in the same frozen evaluator.}
\centering
\setlength{\tabcolsep}{4.0pt}
\renewcommand{\arraystretch}{0.95}
\begin{tabular}{@{}lrrrr@{}}
\toprule
Forward Model
& Eval. CD $\downarrow$
& Eval. TE $\downarrow$
& $|\Delta_{\mathrm{self}}|$ $\downarrow$
& Success $\uparrow$ \\
\midrule
PhysTwin~\cite{jiang2025phystwin}
& 0.0692 & 0.1346 & 0.0322 & 36.4\% \\
EgoPhys \textbf{(Ours)}
& \textbf{0.0565} & \textbf{0.1192} & \textbf{0.0188}
& \textbf{42.4\%} \\
\bottomrule
\end{tabular}
\label{tab:planning_transfer}
\end{table}
Tab.~\ref{tab:planning_transfer} shows that actions planned with EgoPhys achieve lower geometry and tracking error, a smaller gap between predicted and evaluated
outcomes, and higher task success than those planned with PhysTwin. This provides controlled evidence for improved predictive-dynamics action selection under a frozen evaluation model. As a qualitative proof of concept, we execute representative EgoPhys-planned trajectories on two robot arms, an xArm6 and a UR3e, without re-identifying the physical model. We register the simulator and robot frames, convert the planned motions to Cartesian end-effector waypoints, and execute them open-loop with task-specific approach and contact motions. As shown in Figs.~\ref{fig:teaser} and~\ref{fig:robot}, the resulting executions move the deformable objects toward their target configurations. 
\section{CONCLUSION}
We present EgoPhys, a framework for building predictive deformable physical twins from a single egocentric RGB interaction. EgoPhys converts moving-camera monocular observations into a sequence-specific spring--mass model using persistent correspondences, monocular 3D estimates, and egocentric-specific handling of manipulation coupling, support contact, and thin-object dynamics. 
An amortized residual refines local stiffness without dense per-spring test-time optimization, improving held-out prediction. The inferred twins also support action selection, with representative robot executions as a qualitative proof of concept.







\section*{ACKNOWLEDGMENT}

The authors thank the Aria team at Meta and Ramani Duraiswami for their support on hardware. We also thank the Maryland Robotics Center (MRC) and Ivan Penskiy for their assistance in providing access to manipulators.



\bibliographystyle{IEEEtran}
\bibliography{main}

\end{document}